\documentclass{article}

\usepackage[final]{neurips_2024}
\setcitestyle{square}

\usepackage[utf8]{inputenc} 
\usepackage[T1]{fontenc}    
\usepackage{hyperref}       
\usepackage{url}            
\usepackage{booktabs}       
\usepackage{amsfonts}       
\usepackage{nicefrac}       
\usepackage{microtype}      
\usepackage{xcolor}         
\usepackage{graphicx}
\usepackage{makecell}
\usepackage{algorithm}
\usepackage{algpseudocode}
\usepackage{listings}
\usepackage{tabularx}
\usepackage{ltablex} 
\usepackage{xstring}
\usepackage{float}
\usepackage{wrapfig}

\usepackage{abstract}
\usepackage{amsmath}
\usepackage{subcaption}

\title{Augmenting In-Context-Learning in LLMs via Automatic Data Labeling and Refinement}

\author{Joseph Shtok, Amit Alfassy, Foad Abo Dahood, Eliyahu Schwartz, Sivan Doveh and Assaf Arbelle \\  Multimodal AI team, IBM Research AI  \thanks{corresponding author: Joseph Shtok, josephs@il.ibm.com} \\}

\begin{document}

\maketitle

\begin{abstract}

It has been shown that Large Language Models' (LLMs) performance can be improved for many tasks using Chain of Thought (CoT) or In-Context Learning (ICL), which involve demonstrating the steps needed to solve a task using a few examples.
However, while datasets with input-output pairs are relatively easy to produce, providing demonstrations which include intermediate steps requires cumbersome manual work.
These steps may be executable programs, as in agentic flows, or step-by-step reasoning as in CoT.
In this work, we propose Automatic Data Labeling and Refinement (ADLR), a method to automatically generate and filter demonstrations which include the above intermediate steps, starting from a small seed of manually crafted examples.
We demonstrate the advantage of ADLR in code-based table QA and mathematical reasoning, achieving up to a 5.5\% gain.
The code implementing our method is provided in the Supplementary material and will be made available.

\end{abstract}

\begin{wrapfigure}{r}{0.4\textwidth}

\vspace{-14mm}
  \begin{center}
    \includegraphics[width=0.36\textwidth]{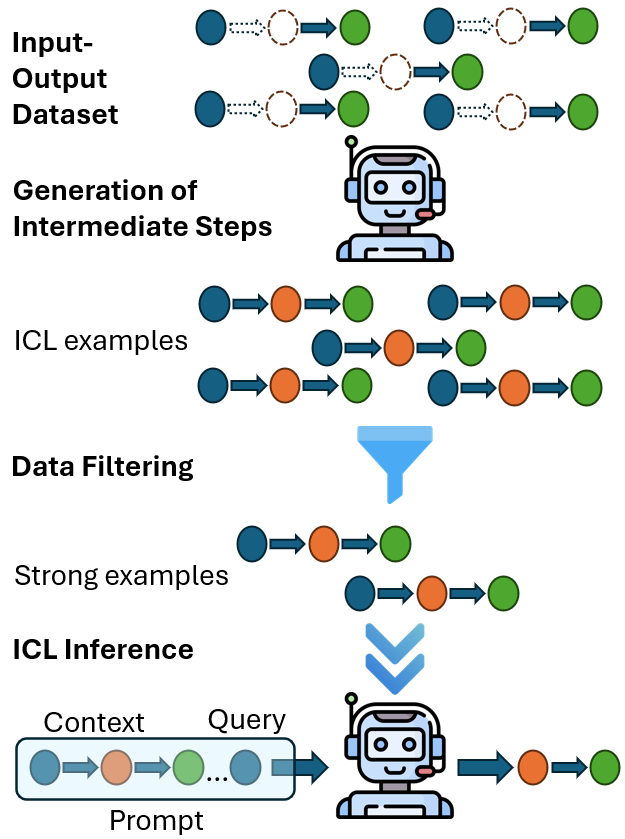}
  \end{center}
  \caption{From an input-output dataset with no intermediate steps (CoT/Executable programs), ADLR generates examples with such steps and retains the useful examples. These examples are used in the ICL to answer a given query.}\label{fig:teaser}
  \vspace{-4mm}
\end{wrapfigure}
\section{Introduction}\label{sect:Intro}
The past decade has seen a big renaissance in the Machine Learning (ML) domain with the rise of neural networks which continue to break all limits at a rapid pace. Until recently, the common training paradigm was based on task-specific models, each trained on a separate dataset for a given task, e.g classification \citep{ref_cls}, detection \citep{ref_detection}, summarizing \citep{ref_summarization}, translation \citep{ref_translation}, etc. 
Today, we see the rise of Foundation Models \citep{ref_foundation} largely based on Large Language Models (LLMs), which have several interesting emerging properties, including  In-Context-Learning (ICL) and Chain-of-Thought (CoT) inference. 
ICL is an approach where the model's behavior is modulated through the model's input, i.e. the context. This context can include information that is required to answer a desired query. This concept is extremely useful in several pipelines, for example in Retrieval-Augmented Generation (RAG)~\citep{ref_RAG} systems. In other cases, the context can include several examples of input-output pairs that outline the models' expected behavior. This property of the model of receiving the small training data for the required task \textit{inline}, without modification of weights, was envisioned at the earlier stage of ML development as the concept of meta-learning \citep{Vinyals2016} and implemented as few-shot learning, where few training examples are appended to the input query.

The ICL regime has proven advantageous for LLMs on general tasks \citep{good_gpt3}, but it is particularly strong when a task, assigned to the LLM, has a particular structure and presumes a specific form of response. In this scenario, the model's input, a.k.a the prompt, consists largely of the context and is composed of several examples exhibiting the domain and format of the input data, as well as the nature of the required output. In fact, these examples contain the expert knowledge that defines the task.
Examples of such special tasks include the CoT approach to solve complex tasks that require reasoning, such as Math Word Problems, with demonstrations of step-by-step mathematical arguments. 
Another example where ICL is successfully used is the neuro-symbolic, or agentic flow - approach which leverages code models to generate executable scripts, which in turn are executed to generate the desired answer~\citep{visprog, viper}. \\
One example is the Table QA task where answering the question requires complex logical and mathematical operations. When tackling this problem with ICL, the LLM's prompt is designed to steer the network into solving the query through task-specific code, usually in Python or SQL. To that end, the prompt examples are designed to contain the table, the question, and the program that should run on the table to answer it. 
The aforementioned expert-knowledge examples are usually hand-crafted for the specific dataset in small amounts, and the same prompt, comprising of those examples, is used for all queries. This approach doesn't leverage the large available training sets, containing diverse samples that can be used despite not being ready-made for the context.

This paper introduces a simple and useful method for improving the performance of existing algorithms operating LLMs in the ICL regime where intermediate steps, such as code generation or CoT is used.  Specifically, we seek to augment the initial set of manually crafted few-shot examples with a much larger set of automatically computed examples which are shown to be (1) hard for the LLM to solve and (2) useful in solving other hard samples in the few-shot context. 

There are then three steps we take towards our goal, as outlined in Figure~\ref{fig:teaser}.
First, given a training dataset comprised of inputs and final answers, we execute the given algorithm using the initial hand-crafted context to generate the intermediate data for these samples. It's correctness is verified by checking that it leads to the correct final answer (by continuing the generation, in case of CoT, or by executing the computer program, in case of code-based QA). This provides us both with a collection of full solved examples, and with a set of hard (unsolved) samples.

In the second step, we refine the collection of solved examples, using the two aforementioned criteria. The difficulty of a sample is estimated by the portion of LLM runs, with the same input and a non-zero temperature, that results in a correct solution.
We restrict the pool of examples to the hard ones, as those possibly contain more new information for the LLM. Further, we subject these examples to the test of utility, seeking to choose those that have proven themselves able to solve many hard samples in a one-shot prompt. 
Finally, we employ the refined pool of examples to augment the inference protocol of the baseline algorithm. We use multiple diverse contexts, populated with random subsets of the pool, and containing a high number of examples. The results of the corresponding multiple runs of the LLM are aggregated via the majority vote (by the common practice).

The contributions of this paper are the following: \textbf{[1]} We propose a method for improved context generation in ICL-based LLM inference for complex tasks improving the current results by up to $5.5\%$;
\textbf{[2]} We conduct an ablation study exhibiting different aspects of the ICL inference as well as those of our method; 
\textbf{[3]} We release an open-source code repository 
to make the method directly accessible by the community (please see details in Supplementary Section \ref{sect:sup_reprod}).


\section{Prior Art}\label{sect:prior_art}

\subsection{Automatic prompt selection and generation for the In-Context Learning}\label{sect:prior_art_selection}


There is good evidence that invoking ICL for the few-shot scenario can improve the performance of an LLM on a number of tasks~\citep{Brown_few_shot}. In this approach, the context of the prompt includes the few-shot exemplars that guide the model in solving the task. Both the selection process itself and the construction of the prompt from the selected examples have been a subject of interest for many prior works. 
Initially, the few-shot examples were carefully and manually curated for every task. 
This manual process is neither scalable nor stable, as there is no guarantee that the manually selected examples are optimal for the task.
A different approach is to select the few-shot exemplars from a large pool of samples with the common practice of randomly selecting the samples from the pool~\citep{Brown_few_shot,Binder,ReAcTable}.
However, it has been shown that the ICL performance can be unstable since it is dependent both on the selection and ordering of the context examples~\citep{ex_selection_ICL,Gao2021,good_gpt3}. 
Some methods strive to stabilize the model by dynamically selecting the context examples depending on the given query. One selection method, used in~\citep{good_gpt3}, is a similarity-based K-nearest neighbors (K-NN) selection, where the semantic similarity is computed using the embeddings of a different model, in this case a pre-trained RoBERTa model. 
In a number of other works, dedicated networks are trained to exercise the sample selection either via Reinforcement Learning in \citep{TabMWP} and \citep{RetICL}, or via contrastive learning in \citep{Rubin22}. 
In our work, we refrain from the ambitious goal of training a model to predict the usefulness of shots for LLMs with respect to a particular query; instead, we gather statistics (over a pool of candidate examples) of the usefulness of the examples from direct one-shot evaluations of the LLM. This results in a much simpler output - a collection of generally strong examples, proven to work for difficult queries. This approach has the added benefit of not requiring an execution of an additional model during inference.

Another direction of improving the sample selection is restricting the pool of examples to a high-quality subset, as was done in \citep{Chang23}. Its authors show that curation of the selection pool both stabilizes the ICL performance and improve it. 
Inspired by these considerations, we incorporate two steps of data filtration into our process, as will be detailed further in Section~\ref{sect:our_method}.

Avoiding the use of the small set of manually crafted expert examples was also considered in~\citep{auto_COT}, which addresses the augmentation of Chain of Thought (CoT) for complex reasoning. The authors demonstrate that reasoning chains, generated with an LLM by a corresponding instruction, can successfully replace the hand-made chains, given enough diversity. For the selection of ``good" samples for the prompt, the authors used semantically-based clustering. In contrast to this work, we extend the hand-crafted examples, rather than replace them, by generating \textit{verified} demonstrations, as detailed in Section ~\ref{sect:our_method}.

\subsection{Table QA with code LLMs}
There are two approaches to addressing the limited reasoning and mathematical skills of LLMs for the table QA task. One way is to fine-tune a general-purpose LLM on data involving operations of this kind~\citep{ReasTAP,Tapex,OmniTab,Cabinet}; the other approach is to employ a code-generating LLM in writing an executable program which performs the necessary operations on the table \citep{Lever,Dater,Binder,ReAcTable,chain-of-table}. Our work belongs to this second group.

Since answering questions on tables may require non-trivial mathematical operations such as condition-based filtering, $max()$, $sum()$ and others, LLMs are not always able to directly answer the queries. Therefore, most leading SoTA solutions for table QA were built on strong off-the-shelf code LLMs\footnote{With the exception of the  work~\citep{Cabinet} from ICLR 2024, which achieves SOTA with a finetuned LLM.}, producing the final answer by executing the generated code on the input table.  Binder~\citep{Binder}  introduced the approach of generating programs with embedded API calls to an external expert to resolve the difficulties of interpreting the question in terms of table headers and rows. Lever~\citep{Lever} introduced an orthogonal feature of a trained model verifying and selecting the output programs from multiple executions of the LLM on the same query, to retain those more likely to produce the correct answer. Dater~\citep{Dater} took yet another approach of helping the LLM to interpret the question by breaking down the process into intermediate tasks.  

Chain-of-Table~\citep{chain-of-table} shift the focus of CoT concept from the logic to the data, gradually updating the table in order to help interpreting the question in terms of its content. 
ReAcTable~\citep{ReAcTable} employ the ReAct paradigm \citep{ReAct} to progressively enhance the input data by generating, similarly to Chain-of-Table~\citep{chain-of-table}, intermediate data representations and injecting the results of the code interpreter back to the LLM querying process. Most of these works could benefit from an orthogonal feature of having a stronger pool of few-shot examples, instead of the few hand-crafted ones. 

\section{The Automatic Data Labeling and Refinement
(ADLR) method}\label{sect:our_method}
\begin{figure}[h]
  \centering
   \includegraphics[width=0.95\linewidth]{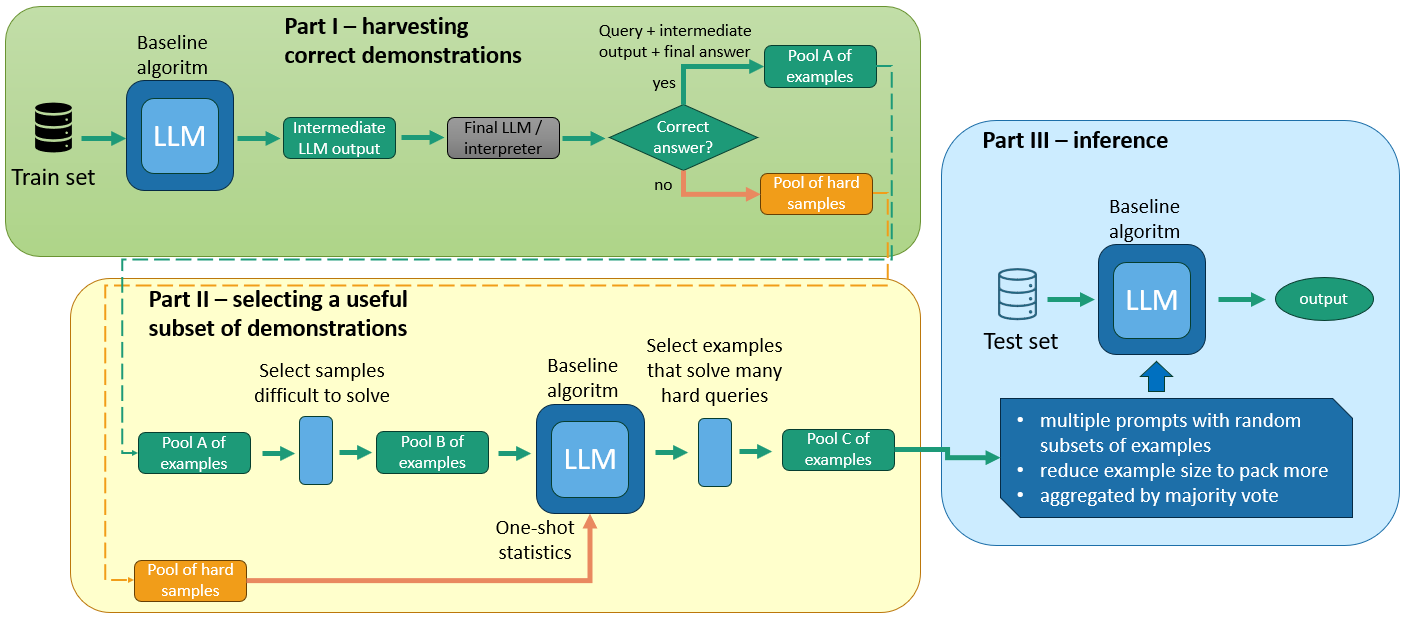}
  \caption{Flow diagram of ADLR}
  \label{fig:diagram}
\end{figure}

As was described in Section~\ref{sect:Intro}, our method consists of three parts (presented in the diagram in Figure \ref{fig:diagram}): 
\textbf{[1]} Produce a large pool of samples annotated with the desired intermediate outputs, which can serve as the few-shot examples (demonstrations);
\textbf{[2]} filter and refine the original pool to contain a subset of high-quality examples; and
\textbf{[3]} perform ICL with the selected examples, using a protocol involving a large set of diversified prompts.

These steps are orthogonal to any particular algorithm using ICL. Parts 1,2 are applied while using the underlying algorithm as a black-box, while the third one requires minimal altering to the algorithm's inference protocol.


\subsection{Part I - generating a large pool of examples}
The first step is to extend the initial hand-crafted set of  examples by using the large training set, that does not have the desired intermediate outputs but does include the final answers. The task is akin to to classical semi-supervised learning (SSL), where there are few annotated samples (in our case, the hand-crafted ones), and a lot of unlabeled. But in contrast to SSL, where there is no possibility to verify the pseudo-labels produced for unlabeled data, we can use the final answers. In the case of Table QA, these are the answers to the table questions, obtained after running the generated code; in the case of CoT, there are the expected final responses of the LLM after going through the chain of thought. 
So we make here the non-rigorous, but statistically plausible assumption that if the intermediate generated data leads to the correct answer, it is by itself correct. Thus we simply accept the intermediate data to build an example, if the final answer is correct. 

By running the inference of the baseline algorithm on the training data, we thus obtain a large pool of fully solved examples - e.g., for WikiTQ, where the performance stands on $60\%-68\%$ accuracy, this yields a pool of examples of size at least $\sim 60\%$ of the training data. We refer to the obtained pool of samples as ``Pool A".

\subsection{Part II - focusing on the useful examples}
On our way to creating strong few-shot prompts, we seek to single out the examples that would help the LLM produce the desired results. This, in general, is a difficult problem, and is approached in the literature by retrieval of examples for a given query (using semantic similarity or a pretrained network, as discussed in  Section \ref{sect:prior_art}). We take a different approach and produce a general set of useful examples, which are randomly sampled to create multiple diverse prompts, as detailed later.

First, we look closer into the LLM inference protocol in tasks like table QA: by common practice the LLM is executed $N$ times (with $N\sim 10$) with the sample prompt, while setting the LLM temperature at non-zero value, and the various answers are aggregated via the majority voting. This technique improves the accuracy and stability of the system \citep{ref_majority}. Also, it enables us to introduce a metric for the level of difficulty of the samples: for a sample $s$, on which $k\leq N$ executions of the LLM have led to the correct result we define the difficulty $D(s)=\frac{k}{N}$. 

\begin{figure}
    \centering
    \includegraphics[width=0.75\linewidth]{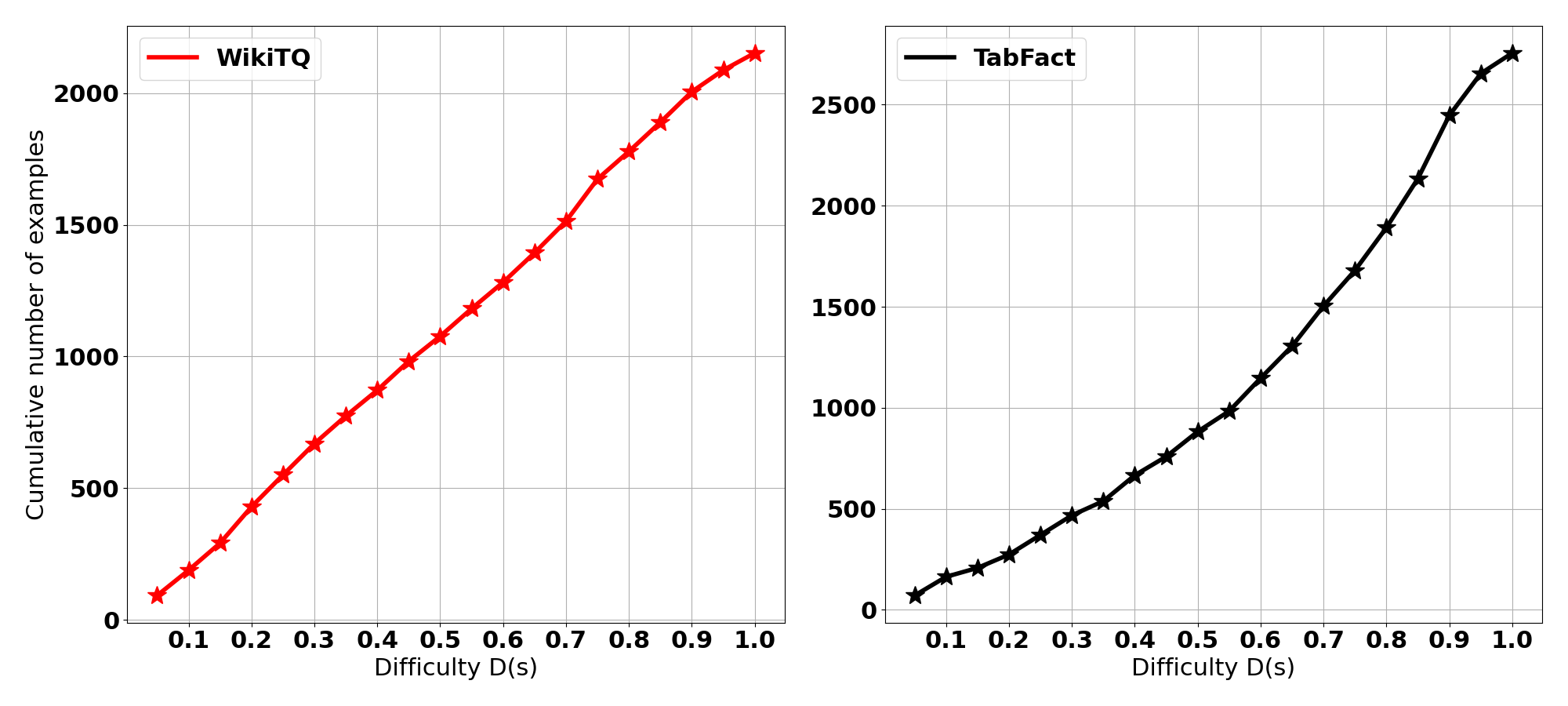}
    \caption{Cumulative distribution of pool-A examples according to the sample difficulty metric $D(s)$}
    \label{fig:stage_B_plots}
\end{figure}

In Figure \ref{fig:stage_B_plots} we plot the number of successfully solved training samples, as a function of the \textit{maximal} number $k$ of attempts the LLM has succeeded for them, for $N=20$ attempts.  Thus, in the case of WikiTQ dataset, out of 3500 used train samples, 92 were solved with just one prompt out of 20, 188 were solved with either 1 or 2 attempts and so on. The takeaway from this analysis is the subset of samples that are, on one hand, solvable, but on the other hand, hard - $D(s)$ restricted from above. We set the threshold at 0.2, using the samples solved with at most 4 out of 20 attempts. These constitute the ``pool B" of our protocol.

The second challenge, posed to our candidate examples, is their usefulness in few-shot scenarios. To gauge that, we run a massive set of experiments with one-shot prompts, populated with examples from pool B, and the queries are selected from the set of samples unsolved at the initial inference on the train data. Interestingly, percentages of those samples that do get solved when a ``right" one-shot prompt is used, are quite high: $68.2\%$ and $71.8\%$ in WikiTQ and TabFact datasets, respectively. This tells us that improving the performance on the given data is well within the capabilities of the LLM, it just needs to receive the right examples in the context.

\begin{figure}
    \centering
    \begin{minipage}{0.45\textwidth}
        \centering
        \includegraphics[width=1.0\textwidth]{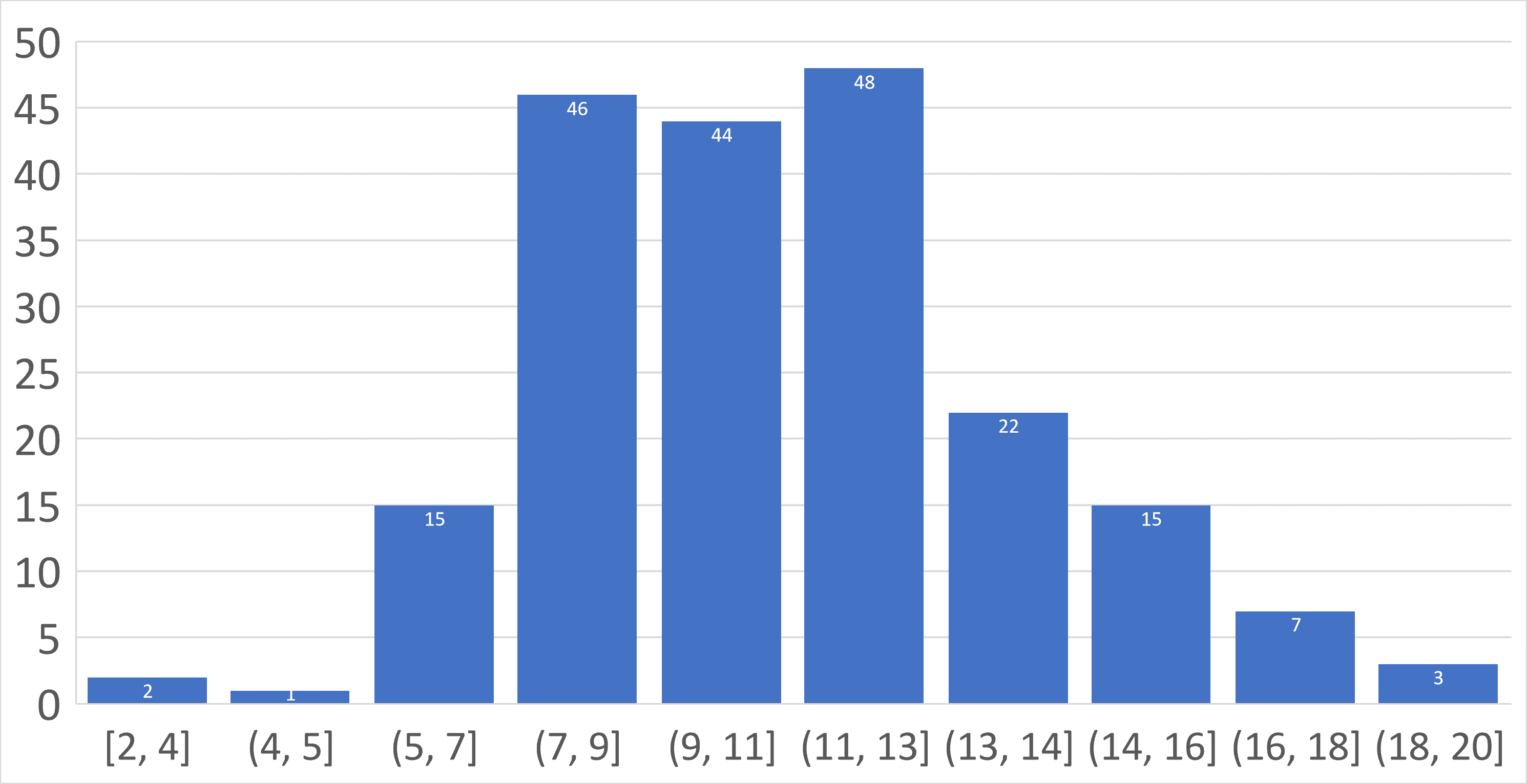} 
    \end{minipage}\hfill
    \begin{minipage}{0.45\textwidth}
        \centering
        \includegraphics[width=1.0\textwidth]{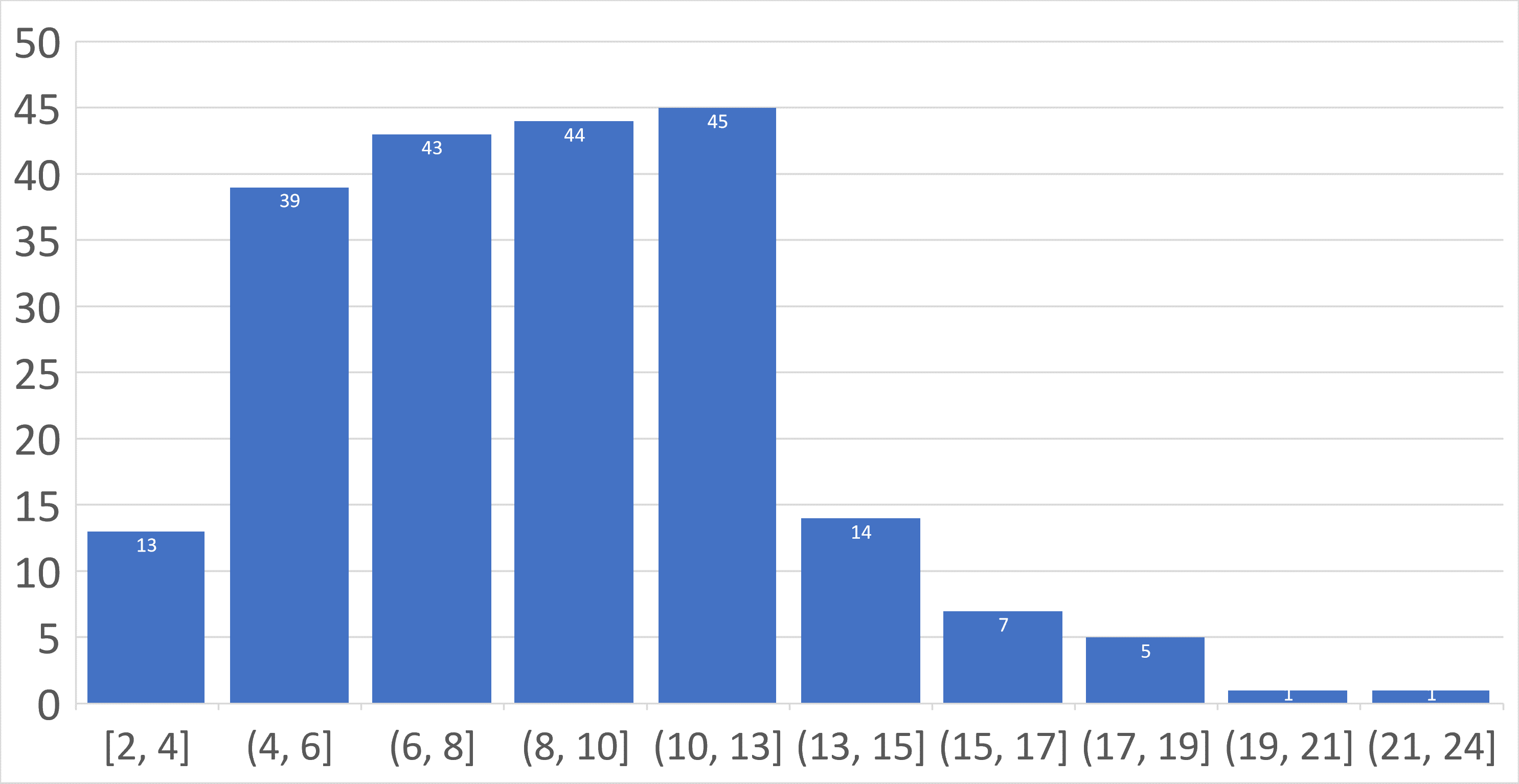} 
    \end{minipage}
        \caption{Histograms of success rate (percents of solved samples) for pool-B examples, applied in one-shot regime}
    \label{fig:shot_stats_hist}
\end{figure}

Now, we have per-example performance: how many of the hard queries was it able to solve. So, we further restrict the pool of examples to those that have solved at least 10\% of the queries. In Figure \ref{fig:shot_stats_hist} we present a histogram of the success rate of the pool B shots, where the $10\%$ threshold was applied. On one hand, the threshold is high enough to retain only the strong examples, on other hand - there is still enough such examples (to the left of the "10" value in the histogram). These constitute the final ``pool C", employed for populating the prompts for ADLR. It is merged with the original, hand-made set of examples.

The work of our pipeline can be visualized via the amounts of samples from the training set that are gradually refined from pool A to pool C. The numbers are given in Table \ref{tab:examples_stats}). 
\begin{table}
    \centering
    \caption{Amounts of examples at different stages of selection, obtained for the Binder implementation}
    \begin{tabular}{ccccc} \toprule
         & \textbf{ Pool A }& \textbf{Pool B }& \textbf{Pool C}& \textbf{ Unsolved Samples} \\ \midrule
         WikiTQ
&  2057&  429&  135& 1433\\ 
         TabFact&  2755&  363&  128& 743\\ \bottomrule
    \end{tabular}

    \label{tab:examples_stats}
\end{table}

\subsection{Part III - augmenting the inference protocol}
With the restricted pool of few hand-crafted examples, inference protocols in most of the relevant algorithms comprise of (1) using a single context, composed of the available examples; (2) running the LLM on the prompt composed of this context and the query multiple times, with non-zero LLM temperature ($0.5-1.0$); (3) integrating the (possibly) different predictions among the successful attempts\footnote{in code-based table QA, some runs may results in a technically incorrect code which fails to run at the interpreter} via a majority vote. This protocol performs better than a single-attempt run \citep{ref_majority}, both because the LLM has a better chance to produce a working code, and because the majority voting helps filtering out outliers.

One of the main features of our approach is the ability to produce multiple diverse contexts, by sampling of the pool of examples. We have experimented with similarity-based K-NN method \cite{good_gpt3}, as detailed in Section \ref{sect:prior_art_selection},  but eventually we chose to use random sampling, with no dependence on the query sample. Beyond that, we used the enlarged examples pool to increase the number of demonstrations in each prompt. This step is, of course, dependent on the size of the examples and the capacity of the employed LLM; in our experiments we used CodeLlama-34b and GPT-3.5-Turbo, that allow for 16k tokens as inputs.

\section{Experiments}\label{sect:exp}

\subsection{Experimental setup}
\textbf{Baseline methods and Datasets}  We evaluate our method on three widely used table QA datasets: WikiTableQuestions (WikiTQ) \citep{WikiTQ}, Table Fact Checking (TabFact) \citep{TabFact}, and the TabMWP \citep{TabMWP}  datasets. 
Since our proposed approach aims to improve the in-context example selection process, we restrict our comparisons to methods that use in-context examples that include intermediate outputs, i.e, either code-generation or CoT.
For the first two datasets, where the task is Table QA, the evaluation is conducted by applying our method to two strong table QA solutions that use intermediate outputs in the form of code generation:  the Binder \citep{Binder} and the ReacTable  \citep{ReAcTable} algorithms. In both cases, we used the original code, shared by the authors, and wrapped it with our algorithm. The final results are presented in Table~\ref{tab:res_tabqa}. On the TabMWP, we demonstrate the merits of our sample selection, continuing the study of different selection policies in the original \citep{TabMWP} paper. Here, ``Part I" of ADLR is not required, since the samples of the training set already contain the detailed ``think step by step" thought process, leading to the answer, a solution of a math problem. However, as can be seen in Table~\ref{tab:tabMWP} the following filtration of the pool of examples proves fruitful for final performance.

\subsubsection{Selecting an LLM}
While Binder has invoked the Codex (code-davinci-002) model by OpenAI, our reproduction (and augmentation) of its algorithm was done with the CodeLlama-34b \citep{ref_codellama}, an open-source model. While the reproduced baseline performance was somewhat lower than that reported in \citep{Binder}, the positive outcome was the availability of a quality open-source solution for these two datasets (just 1.1\% short of the performance attained with the proprietary engine, with our augmentation). 
In the case of ReAcTable, in which the GPT-3.5-Turbo was used, the substitution of LLM with open-source alternative have caused substantially worse results; the large performance gaps due to different LLMs are also observed in the literature \citep{Chameleon}. The reason, we believe, lies in the fact ReAcTable requires more complex outputs, relying more on the LLM's wits. This, in fact, is another hidden factor in choosing the right algorithm for a task solution, not discussed in the current literature: while some methods (like Binder) get somewhat lower performance than those relying on strong LLMs, they may be preferable in the sense that they rely on free models for the text generation and thus reducing the cost of operations.
In light of these findings and since Codex is no longer available, we have reproduced the ReAcTable evaluation using GPT-3.5-Turbo.
Additionally, for TabMWP, the selection of the LLM also plays an important role. As was shown in \citep{Chameleon}, GPT-4 of OpenAI can attain 91.81\% accuracy in the few-shot regime in a plain application to this data (i.e., with the provided CoT inputs), while GPT-3 delivers only 62.92\% in this regime. We again have experimented with some open-source LLMs, showing intermediate results; as our purpose here is to show a relative improvement (for different selection policies) using our proposed algorithm, the results are reported for the CodeLlama-34b \citep{ref_codellama} and the Mixtral-8x7b-instruct \citep{ref_mixtral} models.

\subsubsection{ADLR parameters and inference protocols}
We summarize here the meta-parameters used in applying our method, including the parameters of its inference protocol, as well as the baseline inference protocol.
Part I has no parameters to set - we simply run the baseline algorithm through the training set in order to harvest solved and unsolved samples. For the reasons of computation savings, we used only 3500 samples for each of the WikiTQ and TabFact datasets, and 8k samples from the TabMWP data. The reason for larger pool in TabMWP is that in the inference the samples are further partitioned into question types and answer types, and we need a sufficient amount of examples of each type to populate the contexts. Examples of TabMWP data, as well as details of questions/answers types, are provided in Supplementary material.

In Part II, pool B is produced as a subset of pool A of samples which are solved in at most 20\% of LLM runs: $D(s)=\frac{k}{N}<0.2$. 
Then in the one-shot statistics, each shot (A pool-B member) is selected for pool C if (1) it was used on at least 100 hard samples, and (2) it has solved at least 10\% of these samples.


\subsection{Finetuning an LLM}
One immediate usage for the pool-A data is to finetune the LLM to generate  the desired intermediate outputs (correct programs for table QA, or the thought chains) rather than the final answers, which were used as labels in the prior works so far. This approach aligns with the initial reason we need the intermediate data in the first place - on these tasks, it is challenging for the LLM to generate the final answer directly. We have gathered 4800 train samples for the WikiTQ dataset and 5000 samples for the TabFact, with the Binder algorithm. The CodeLlama-34b model was finetuned using the LoRA extension \citep{ref_lora} with parameters $rank=64,\alpha=16$. The model has trained for 20k steps till convergence. We witness the accuracy improvement by $3.1\%$  on the test data of WikiTQ in Table \ref{tab:res_tabqa}. Additional data is available in the ablation study (Figure \ref{fig:ablation_stages_nshots}). Unfortunately, we were not able to repeat the process for the ReAcTable algorithm, as the LLM used is this case is proprietary.

\subsection{Computational resources}
For inference with Codellama-34b and Mixtral-8x7b-instruct models we have used eight A100 80GB GPUs by NVIDIA, on internal server with 1280GB of RAM memory, loading them in bf16 precision. The same resources were used for the training of Codellama-34b. We should mention that the full research project required more compute than the experiments reported in the paper, in order to develop the method. 

\subsection{Results on Table QA task}
The application of our method to Binder and ReAcTable solutions for the two Table QA datasets yields the results presented in Table \ref{tab:res_tabqa}. Unfortunately, the authors of ReAcTable have only released the data and the few-shot examples for the WikiTQ dataset, thus the results on TabFact are not reported for this algorithm.
Due to using different LLMs, and possibly other factors, we were not able to fully reproduce the accuracy reported in the original papers. However, we have shown the merits of our method by comparing to the reproduced baseline, having all factors equal except the inference protocol. With Binder algorithm, CodeLlama-34b has shown a result close to the reported Codex performance ($58.9\%$ accuracy on WikiTQ, instead of the published $64.6\%$; a similar gap is observed for TabFact). With ReAcTable, the gap is unfortunately much larger ($22.4\%$); this may be attributed to the more complex instructions and the Program-of-Thought chains employed by the algorithm. So one conclusion is - while with a ``smart" enough LLM, ReAcTable achieves better performance, if money is a factor, and using an open-source model is desired (possibly for other reasons, too), Binder is actually better.
Our ADLR enhancement helps increase the accuracy by $2.0\%$ - $2.8\%$.

\begin{table}
    \small
    \centering
    \caption{Results of Binder and ReAcTable on two TableQA datasets, WikiTQ and TabFact. The reproduction was done using the GitHub repositories of the relevant methods. The original results using Codex were not reproducible. We show significant gains due to our algorithm for both methods. $^*$"FT" stands for finetuned LLM.}
    \begin{tabular}{lllcc} 
         \toprule
         \textbf{Algorithm}& \textbf{Comments} &  \textbf{LLM }& \textbf{WikiTQ} & \textbf{TabFact}\\ \toprule
         Binder& published by authors &   Codex&64.6\% & 85.1\% \\ 
         Binder& reproduced by us&   CodeLlama-34b&58.9\%& 78.6\%\\ 
         Binder+ ADLR  & Ours  &  CodeLlama-34b&61.2\%& 80.6\%\\ 
 Binder+ ADLR  & Ours & CodeLlama-34b-FT$^{*}$& 64.3\%&83.8\%\\  
 \midrule
         {ReAcTable}&published by authors &   Codex&68.0\%\\ 
         {ReAcTable}&reproduced by us&   GPT-3.5-Turbo&62.5\%\\ 
        {ReAcTable}&reproduced by us & Code-Llama-34b& 45.6\%\\ 
         {ReAcTable + ADLR }& Ours &   GPT-3.5-Turbo&65.3\%\\ 
         \bottomrule
    \end{tabular}
    
    \label{tab:res_tabqa}
\end{table}


\subsection{Results on the TabMWP dataset}
Following \citep{TabMWP}, we study the performance of LLMs on the TabMWP dataset with different sample selection policies. 
Since the training samples include step-by-step thinking, we interpret the available data as the ``Pool A" and apply Part II of ADLR method to find a good subset of those. Results in Table \ref{tab:tabMWP} exhibit an ablation study on the context size and the number of LLM runs for each sample, and also exhibit the gain by ADLR, compared to basic selection policies. We find it to be $5.5\%$ and $2.6\%$ for the CodeLlama-34b and the Mixtral-8x7b-instruct models, respectively. The accuracy obtained here is indirectly comparable to the results with the PromptPG algorithm \citep{TabMWP} with RL-trained sample selection model. They attain $68.2\%$ accuracy using the GPT-3 model in 2-shot ICL regime.
\begin{table}
    \small
    \centering
    \caption{Results on the TabMWP dataset for various configurations. Q-type / A-type mean restriction to examples which questions / answers of the same type as the query.}
    
    \begin{tabular}{lcccc} \toprule 
         \textbf{Sample Selection}& \textbf{ \# Shots} &\textbf{\# Attempts}&  \textbf{CodeLlama34b}& \textbf{Mixtral-8x7b-instruct}\\ \midrule
 No restriction& 10&5& 65.3\%&76.4\%\\ 
 Q-type +A-type& 10& 5& 67.3\%&83.3\%\\ 
 Q-type +A-type& 10& 11& 70.1\%&84.4\%\\
         Q-type +A-type&  20&11&  72.1\%& 83.9\%\\ 
          Q-type +A-type +ADLR&  20&11&  77.6\%& 86.5\%\\ 
         \bottomrule
    \end{tabular}
    
    \label{tab:tabMWP}
\end{table}

\begin{figure}[h]
    \centering
    \begin{minipage}{0.45\textwidth}
        \centering
        \includegraphics[width=0.95\textwidth]{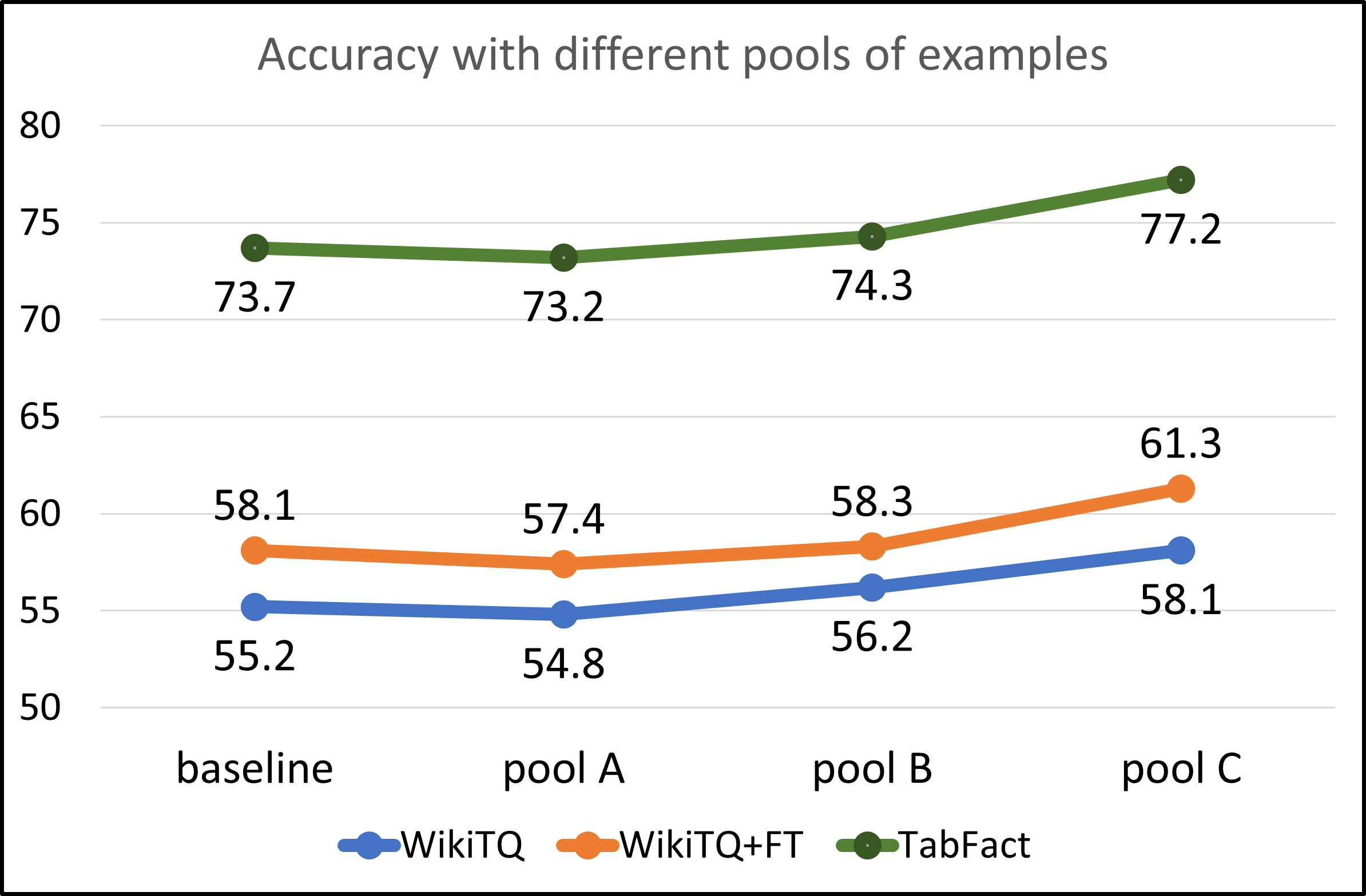} 
        
    \end{minipage}\hfill
    \begin{minipage}{0.45\textwidth}
        \centering
        \includegraphics[width=0.95\textwidth]{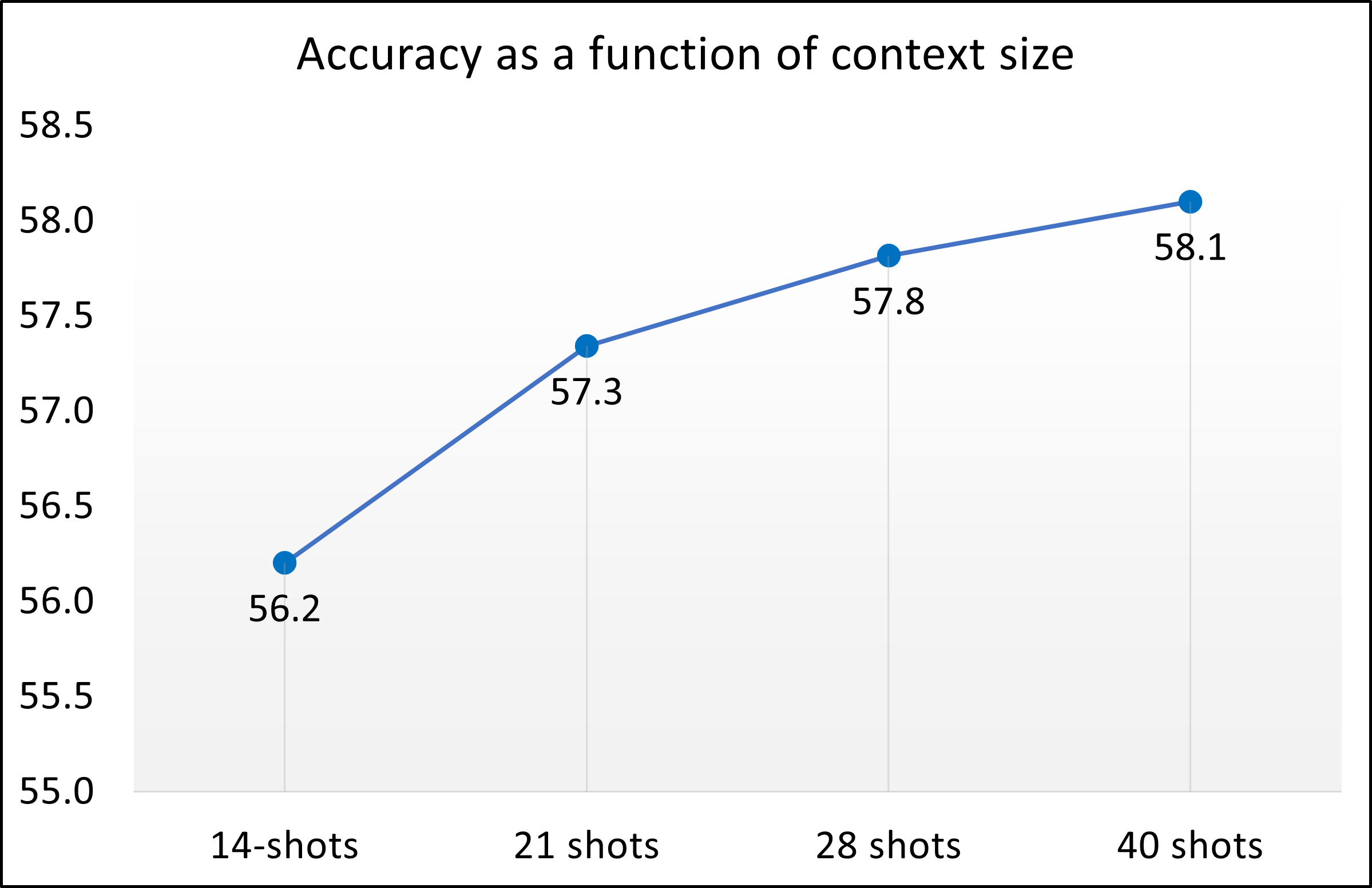} 
        \label{fig:abl_nshots}
    \end{minipage}
        \caption{\textbf{Left:} performance with different pools of examples. For WikiTQ dataset, we compare the vanilla and the finetuned CodeLlama-34b models (WikiTQ and WikiTQ-FT in the legend).  \textbf{Right:} Performance as a number of shots for ADLR, compared to the baseline Binder (fixed context)}
         \label{fig:ablation_stages_nshots}
\end{figure}
\section{Ablation study}\label{sect:ablation}
We now study different aspects of our algorithm, as well as the parameters of inference protocols ofr LLMs in general. In this study, we use the baseline algorithm where the demonstrations contain plain programs code as the requested response, without the Neural API of Binder or the recurrent application of ReAcTAble.

\textbf{Refinement phases of context examples }
In Figure \ref{fig:ablation_stages_nshots} (left) we present the performance levels obtained with different pools of the harvested examples. The baseline computation was done with the 14 plain-code hand-crafted examples provided by Binder authors. We observe that using the entire pool A actually decreases the performance, but restricting it to the hard (interesting) examples, pool B, and further to one-shot useful examples, pool C, helps gaining a significant advantage. Additionally, the accuracy is computed with our finetuned version of the CodeLlama-34b; it shows improvement of $2-3\%$ across the pools. The baseline pool is the original 14 hand-crafted examples provided by authors of Binder \citep{Binder}. In all experiments, we use 20 runs of the LLM per sample. 

\textbf{Context size} One of the features unlocked by the ADLR is to enlarge the context size, originally constrained by the number of manually prepared examples. In table \ref{tab:tabMWP} the ablation study of the \#~Shots parameter has shown mixed results - using a context of size 20 instead of 10 improved the performance for CodeLlama-34b, but - interestingly - decreased the accuracy of the (stronger) Mixtral model. 
In the right part of Figure  \ref{fig:ablation_stages_nshots} we present accuracy as a function of \#~Shots on  WikiTQ. Here the performance improves, as expected. The last step of passing from 28 to 40 shots, while maintaining 16k tokens restriction of the input length, was possible thanks to using a more compact representation of the table in the query (we provide details in the Section \ref{sect:compact} of Supplementary material).

\textbf{Similarity-based shots selection}\label{sect:ablation_protocol} Following \citep{good_gpt3}, we have also attempted to use adaptive retrieval of the few-shot samples using semantic similarity, as was detailed in Section \ref {sect:sup_distr} 
However, this method didn't prove useful in our case. First, it comes at cost of using a single context (with K-NN-based shots selection)  rather that multiple different ones; second, from the one-shots experiments we had the ground truth about which shots are the ``correct" ones for a given query (those that solve it in one-shot prompt). It allows us to observe the distribution of such correct shots in the list of all shots, ranked by the similarity. Unfortunately, we don't see the desired bias of the ``good" shots towards the head of the list, which means this metric correlates poorly with usefulness of the shots. More details on this study are given in the Supplementary material Section \ref{sect:sup_distr}. Quantitatively, executing the single K-NN based prompt 20 times with LLM temperature $0.5$ yields accuracy of $58.3\%$ on WikiTQ, compared to the $61.2\%$ obtained with 20 randomly samples prompts; similarly, on TabFact the performance decreases from $80.6\%$ to $75.9\%$. We conclude that in our case the K-NN method is suboptimal.


\section{Afterword}\label{sect:summary}
\paragraph{Summary and future work}
In this work, we have presented a generic method for augmentation of In-Context Learning of LLMs, for the scenario where hand-crafted demonstrations are required. The method is independent of the given task and algorithm for its solution, and seeks to better exploit the weakly labeled data available for the task. This method can be easily adapted to enhance and improve both proprietary and, especially, open source models, where each incremental enhancement is beneficial and crucial, particularly in environments that rely on existing data and hardware. Further exploration of this approach for other tasks, requiring intermediate data, may help us in developing the method and make it more general, thereby inspiring continuous innovation and collaboration within the community.


\paragraph{Limitations}
The method applies to a specific domain of LLM applications, namely those requiring generation of the intermediate data, such as CoT and executable code programs.

\paragraph{Potential positive and negative societal impacts of our work}
This work promotes usage of large language models (LLMs) in context of complex question answering. While in general this technology is considered to have positive impact in the society by providing useful service satisfying many needs, it can also have a negative side, if used for producing false textual or visual information. Bias Amplification and Privacy are the main concerns related to LLMs; the usage of our method is therefore conditioned on using well curated and legal datasets.

\clearpage
\bibliographystyle{ACM-Reference-Format}
\bibliography{references.bib}

\appendix
\clearpage
\setcounter{page}{1}



\section{Reproducibility}\label{sect:sup_reprod}
ADLR code is based on Open Source (OS) releases of the BInder and ReAcTable codebases. We submit the code representing a fork of the \hyperlink{https://github.com/yunjiazhang/ReAcTable}{github repo} for the ReAcTable algorithm, with some modifications and additional files implementing our method. Our code will be released as an Open Source github contribution upon acceptance.

\section{Evaluation of the K-NN sample selection based on semantic similarity}\label{sect:sup_distr}
As mentioned in Section \ref{sect:ablation_protocol}, we have studied the relevance of the semantic similarity (of the candidate examples to the query), to the success of the LLM on this query when using these examples in the context. Specifically, we used the model \textbf{all-mpnet-base-v2} of type SentenceTransformers \citep{sentense_BERT} to compute embeddings of the samples, where each sample is represented by the text string of the form [(table headers):(one table row):(question)]. The embeddings are unit-norm vectors, so the compute the cosine similarity between the query and each pool example to get ranking of the pool with respect to the given query.

\begin{figure}[h]
    \centering
    \includegraphics[width=1\linewidth]{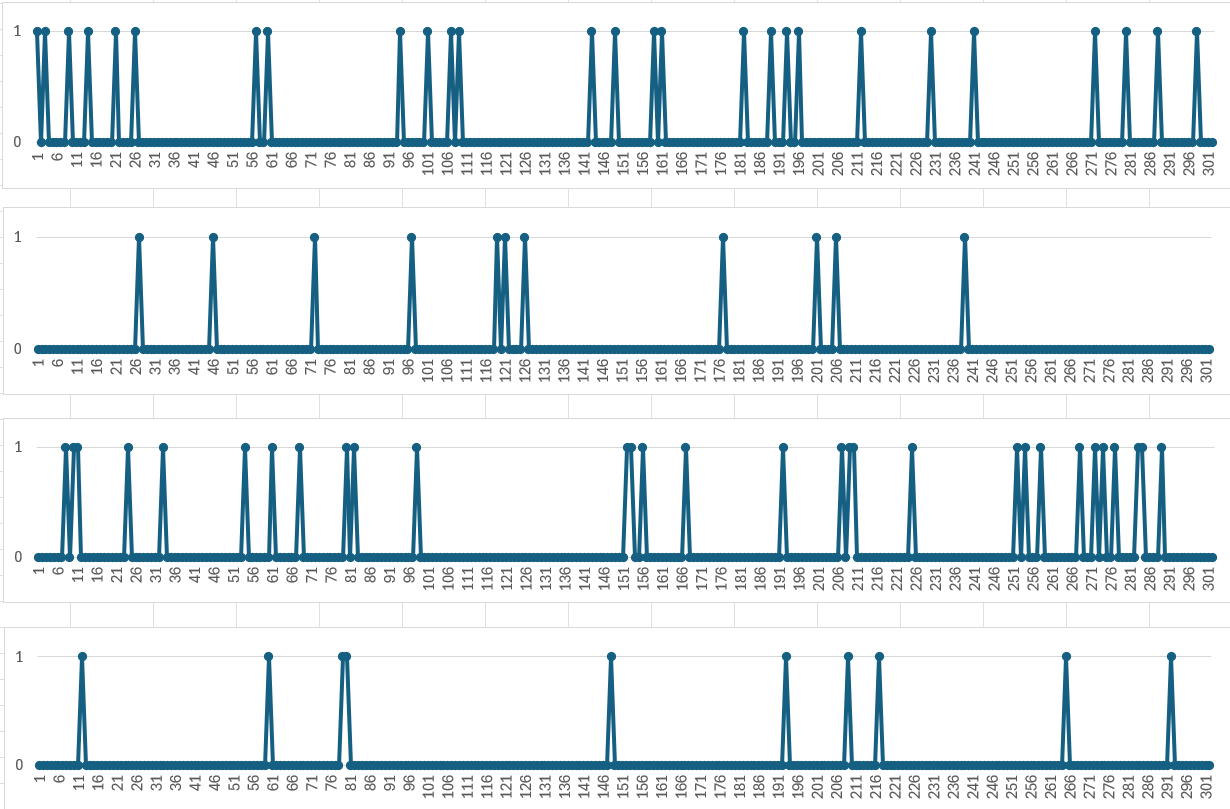}
    \caption{Distributions of the good shots in the similarity-ranked list of the pool examples, for a few queries}
    \label{fig:distr}
\end{figure}

In second stage of part II of our method, we are executing one-shot inference, where the shots come from pool B of solved examples, and the queries are from the set of difficult samples. Thus for each query sample we get the list of shots which helped the LLM solve it (the "good shots"). Now, combining this information with the ranking of the same pool of shots by the semantic similarity, we obtain the distribution of the good shots in the entire pool; we plot the obtained distributions for random four queries in Figure \ref{fig:distr}. Unfortunately, the desired bias of the distributions towards the head of the list (left) is not observed; this accounts for the fact the performance on the test set was not improved by using the K-NN based approach.

\section{Data and context examples for the relevant datasets and algorithms}\label{sup:data&context}
\subsection{WikiTQ - example of automatic annotations}
\begin{figure}
    \centering
    \includegraphics[width=1\linewidth]{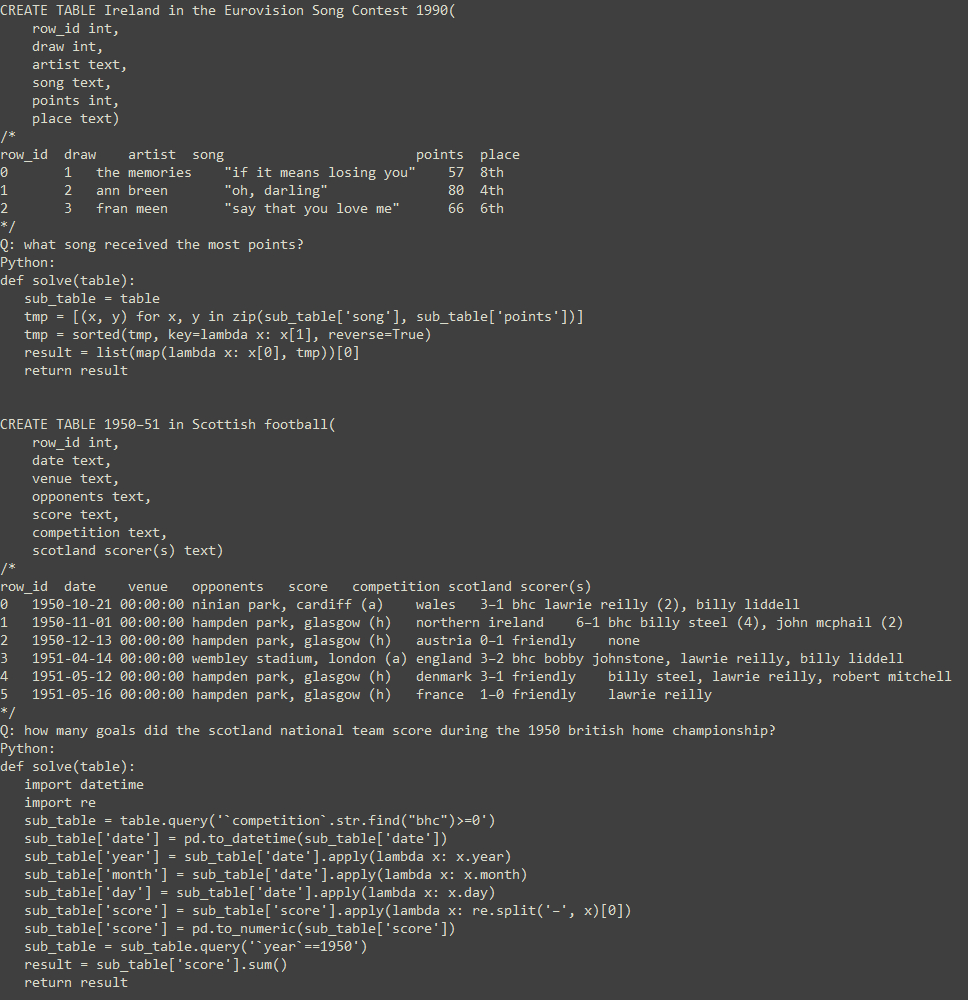}
    \caption{Examples of train samples with automatically generated code}
    \label{fig:wikitq_examples_A}
\end{figure}

We provide some examples of the train samples with automatically generated python code required for the context demonstrations. These are given in Figure \ref{fig:wikitq_examples_A}. The text is structured by the \texttt{\textbackslash t,\textbackslash n} markers, so it is actually given as a single prompt-ready text string. In the displayed examples one can learn about the coding style of CodeLlama-34b, albeit influenced (strongly) by the original hand-made examples. In the second example, the operations are broken into small steps, with some repeatability a human programmer might replace with a loop over different headers.



\subsection{WikiTQ - Compact table representation in context}\label{sect:compact}
\begin{figure}[H]

\subfloat{%
    \includegraphics[width=1.0\linewidth]{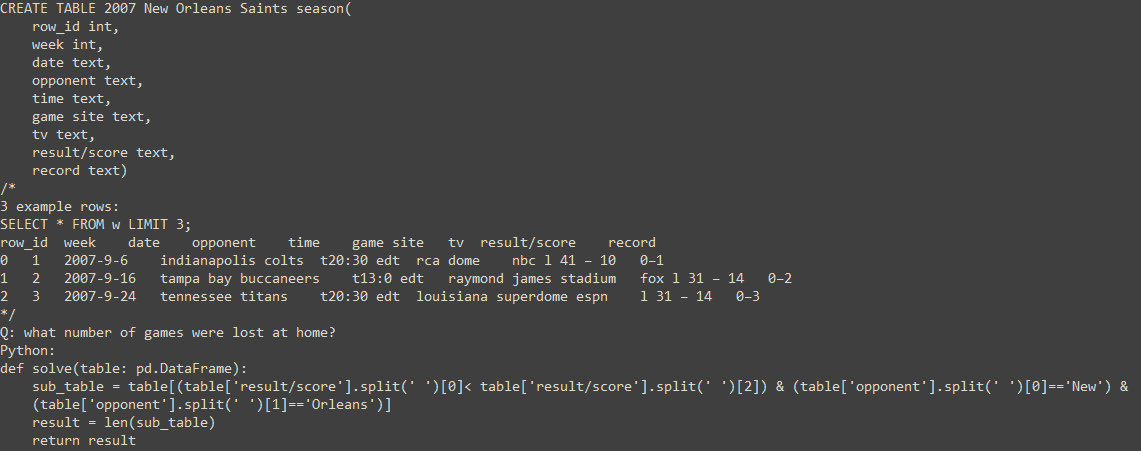}
}

\subfloat{%
    \includegraphics[width=1.0\linewidth]{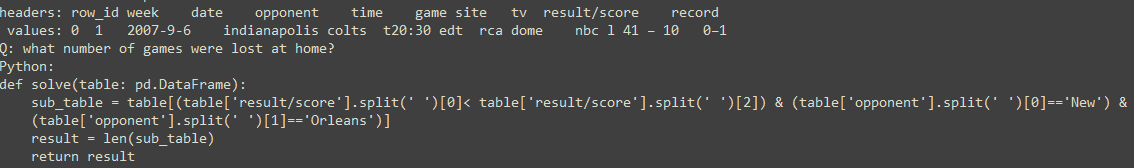}
}
\caption{Top: full context example from WikiTQ dataset, in Binder format; bottom: the same example in our compact format }
\label{fig:short_shots}
\end{figure}

We have found it beneficial to enlarge the number of shots in the context, even at expense of reducing the size of each example to fit into the prompt size restriction (and also because LLMs loose track of information in very long prompts). Specifically, we begin with the format in which the context examples were presented by the Binder authors (Figure \ref{fig:short_shots}. Specifically, we have [1] removed the table header and the column types, [2] removed the \text{/*, ..., */} brackets and the \text{"3 example rows: SELECT * FROM w LIMIT 3"} text, repeating in every example, and [3] retained only one table row instead of three. The question and the desired output (program) were retained in full.

\subsection{TabMWP}
The training part of the TabMWP dataset contains 23k samples, each representing a table, a question (mathematical problem) involving the table data, a reasoning step-by-step process (text), and finally the answer. Each samples is classified according to a question type, answer type and grade, as detailed in Table \ref{tab:tabMWP_types}. 

\begin{table}[H]
    \centering
    \caption{Types of questions and answers in the tabMWP dataset}
    \begin{tabular}{lll} \toprule
         &  \textbf{Name}& \textbf{Amount in training data}\\ \midrule
         Question type&  free\_text
& 17315\\ 
         Question type&   multi\_choice& 5744\\ 
         Answer type&  integer\_number
& 13804
\\
         Answer type&  extractive\_text
& 2991 
\\
         Answer type&  decimal\_number
& 3511 
\\ 
         Answer type&  boolean\_text
& 2467 
\\ 
         Answer type&  other\_text& 286\\ 
         grades&  1-8& \\ 
\bottomrule
    \end{tabular}
    
    \label{tab:tabMWP_types}
\end{table}

As was shown in the ablation study (Table \ref{tab:tabMWP} in the paper), the LLM benefits from receiving in-context examples with question/answer of same type as a query. A few examples of the data are provided in Table \ref{tab:sup_tabMWP_examples}:
\begin{table}[H]
\caption{Examples from the tabMWP dataset}
    \centering
    \begin{tabularx}{\linewidth}{p{0.2\linewidth} | p{0.25\linewidth} | p{0.35\linewidth} | p{0.07\linewidth} | p{0.06\linewidth}}
    \toprule
    Table & Question & Solution & Choices & Answer \\
    \midrule
    \small
\begin{tabular}{cc}
Stem   & Leaf \\
\hline
3   & 3, 3, 3, 5, 5 \\
4   & 6 \\
5   & 4, 5, 7, 8 \\
6   & 7, 8\\
7   & 2, 3, 7, 9\\
8   & 6, 8, 9 
\end{tabular}
    &
    The members of the local garden club tallied the number of plants in each person's garden. How many gardens have at least 47 plants?
    &
    Find the row with stem 4. Count all the leaves greater than or equal to 7. Count all the leaves in the rows with stems 5, 6, 7, and 8. You counted 13 leaves, which are blue in the stem-and-leaf plots above. 13 gardens have at least 47 plants.
    &
    NA
    &
    13\\
    \midrule
    \small
    \begin{tabular}{cc}
    Day & Tickets \\
    \hline
Friday 	& 71 \\
Saturday & 74 \\
Sunday & 75 \\
Monday & 72 \\
    \end{tabular}     
    & 
    The transportation company tracked the number of train tickets sold in the past 4 days. On which day were the fewest train tickets sold?
    &
    Find the least number in the table. Remember to compare the numbers starting with the highest place value. The least number is 71. Now find the corresponding day. Friday corresponds to 71.
    &
    Friday, Saturday, Sunday, Monday
    &
    Friday
    \\
    \midrule
    \small
    \begin{tabular}{cc}
Level & Number\\
\hline
Gold & 15 \\
Silver & 68 \\
Bronze & 58
    \end{tabular}     
    & 
    The Burlington Symphony categorizes its donors as gold, silver, or bronze depending on the amount donated. What fraction of donors are at the bronze level?
    &
    Find how many donors are at the bronze level. 58. Find how many donors there are in total. 15 + 68 + 58 = 141. Divide 58 by 141. 58/141. 58/141 of donors are at the bronze level.
    &
    NA
    & 
    58/141 \\
    \bottomrule
    \end{tabularx}
    
    \label{tab:sup_tabMWP_examples}
\end{table}
\end{document}